\documentclass[runningheads]{llncs}

\usepackage{eccv}

\usepackage{eccvabbrv}

\usepackage{graphicx}
\usepackage{booktabs}
\usepackage{array}
\usepackage{epsfig}
\usepackage{amsmath}
\usepackage{amssymb}
\usepackage{bm}
\usepackage{bbding}
\usepackage{arydshln}
\usepackage{float}
\usepackage{pifont}

\newcommand{\xmark}{\ding{55}}%
\usepackage{listings}
\usepackage{dsfont}
\usepackage{color}
\usepackage{algorithm}
\usepackage{algpseudocode}
\usepackage{multirow} 
\usepackage{tikz}
\usepackage[accsupp]{axessibility}  
\newcommand*\repeatthanks[1][\value{footnote}]{\footnotemark[#1]}
\makeatletter
\renewcommand*{\@fnsymbol}[1]{\ensuremath{\ifcase#1\or *\or \dagger\or \ddagger\or
   \mathsection\or \mathparagraph\or \|\or **\or \dagger\dagger
   \or \ddagger\ddagger \else\@ctrerr\fi}}
\makeatother

\usepackage{hyperref}

\usepackage{orcidlink}

\begin{document}

\title{Betrayed by Attention: A Simple yet Effective Approach for Self-supervised Video Object Segmentation} 

\titlerunning{Betrayed by Attention}


\author{Shuangrui Ding\inst{1}\thanks{Equal Contribution}\orcidlink{0000-0001-7033-774X} \and
Rui Qian\inst{1}\repeatthanks\orcidlink{0000-0002-0378-6438} \and \\
Haohang Xu\inst{2}\orcidlink{0000-0002-4715-1338
} \and
Dahua Lin\inst{1,3}\thanks{Corresponding author. Email: dhlin@ie.cuhk.edu.hk}\orcidlink{0000-0002-8865-7896} \and
Hongkai Xiong\inst{2}\orcidlink{0000-0003-4552-0029}}
\authorrunning{S.~Ding et al.}

\institute{The Chinese University of Hong Kong, Hong Kong, China \and
Shanghai Jiao Tong University, Shanghai, China \and
Shanghai Artificial Intelligence Laboratory, Shanghai, China
\email{\{ds023,qr021\}@ie.cuhk.edu.hk} 
}

\maketitle

\begin{abstract}
    In this paper, we propose a simple yet effective approach for self-supervised video object segmentation (VOS). Previous self-supervised VOS techniques majorly resort to auxiliary modalities or utilize iterative slot attention to assist in object discovery, which restricts their general applicability. To deal with these challenges, we develop a simplified architecture that capitalizes on the emerging objectness from DINO-pretrained Transformers, bypassing the need for additional modalities or slot attention. Our key insight is that the inherent structural dependencies present in DINO-pretrained Transformers can be leveraged to establish robust spatio-temporal correspondences in videos. Furthermore, simple clustering on this correspondence cue is sufficient to yield competitive segmentation results. Specifically, we first introduce a single spatio-temporal Transformer block to process the frame-wise DINO features and establish spatio-temporal dependencies in the form of self-attention. Subsequently, utilizing these attention maps, we implement hierarchical clustering to generate object segmentation masks. To train the spatio-temporal block in a fully self-supervised manner, we employ semantic and dynamic motion consistency coupled with entropy normalization. Our method demonstrates state-of-the-art performance across three multi-object video segmentation tasks. Specifically, we achieve over 5 points of improvement in terms of FG-ARI on complex real-world DAVIS-17-Unsupervised and YouTube-VIS-19 compared to the previous best result. The code and checkpoint are released at \url{https://github.com/shvdiwnkozbw/SSL-UVOS}.
\end{abstract}

\begin{figure*}
    \centering
    \includegraphics[width=\linewidth]{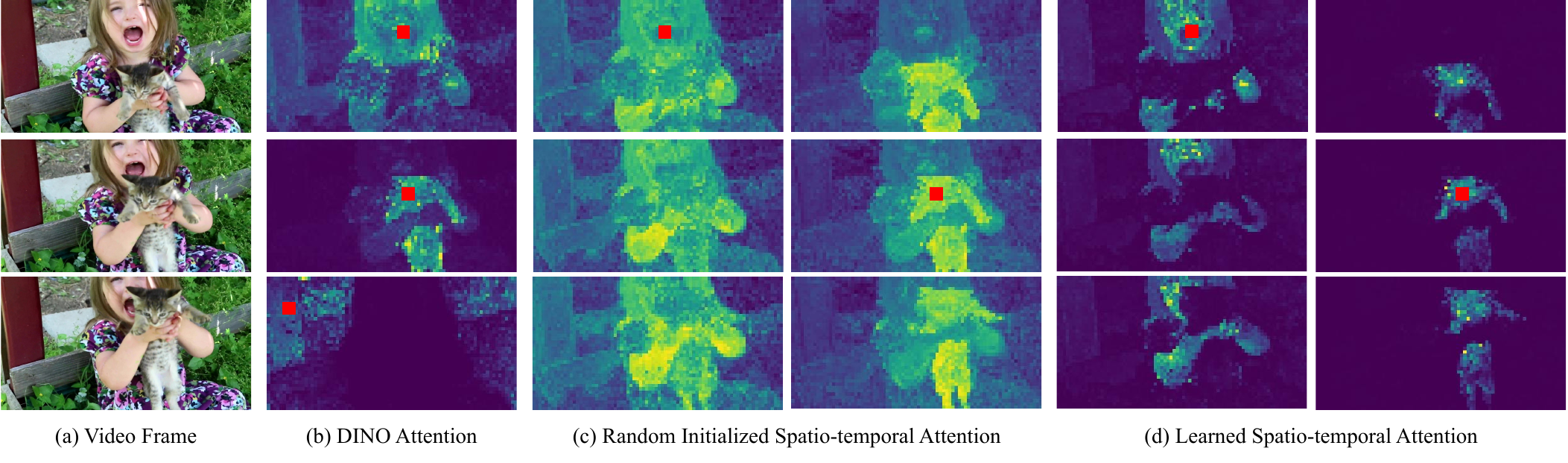}
    \caption{\textbf{Attention leaks the object's position!} We visualize the self-attention maps of different queries (\protect\tikz{\fill[red] (0,0) rectangle (0.2,0.2);} prompt) from the video sequence (a). The frame-wise DINO attention maps (b) highlight image regions corresponding to the queried object. A randomly initialized spatio-temporal Transformer block on top of DINO produces noisy spatio-temporal attention maps (c) that coarsely track objects over time. Our method diminishes noise in the learned spatio-temporal attention maps (d) which derive temporally coherent object segmentation.}
    \label{fig:teaser}
\end{figure*}

\section{Introduction}
\label{sec:intro}
Representing the visual scene with objects as the basic elements has been long acknowledged as a core cognitive capability of an intelligent agent. 
In the realm of computer vision, Video Object Segmentation (VOS) tasks require a model to segment and track specified objects within a video sequence, striving to emulate this foundational function.
This task holds significant importance in various real-world vision systems, including but not limited to, autonomous driving~\cite{feng2020deep} and surveillance security~\cite{patil2021unified}. 
However, traditional methods for VOS~\cite{caelles2017one, perazzi2017learning} typically follow a fully supervised paradigm and entail substantial costs for obtaining pixel-level per-frame annotations. This factor significantly limits their feasibility for large-scale applications.
Thus, the researchers turn their attention to self-supervised learning~\cite{he2020momentum, chen2020simple, qian2021enhancing, qian2022static, ding2021motion, ding2022dual, xu2022masked,chen2022sdae, Liu_2023_CVPR}, a more efficient approach that leverages unlabeled data.

Despite its promising potential, self-supervised VOS poses significant challenges. 
Firstly, previous methods tend to incorporate additional signals such as optical flow~\cite{Yang_2019_CVPR, lian2023bootstrapping, Yang_2021_ICCV}, or depth cues~\cite{kipf2022conditional, elsayed2022savi++}, to furnish object-related clues.
However, these auxiliary signals within videos can often be difficult to extract and unreliable when it comes to multi-object scenarios, thus limiting their availability.
Secondly, most existing models~\cite{qian2023semantics, aydemir2023self, zadaianchuk2023objectcentric} adopt slot attention~\cite{locatello2020object} to parse the frame into objects.
Nevertheless, the requirement for a predefined quantity of learnable slot queries limits its adaptability to in-the-wild data, especially when the number of objects is unknown.
To address these limitations, we recall the recent self-supervised learning technique DINO~\cite{caron2021emerging, oquab2023dinov2}, which demonstrates emerging objectness in the attention maps of pretrained Vision Transformer (ViT)~\cite{dosovitskiy2020image}.
As depicted in Fig.~\ref{fig:teaser}(b), DINO attention maps encode spatial dependencies between diverse patches. These dependencies present different patterns for various objects, providing abundant cues for object segmentation. 
This phenomenon motivates us with an intriguing possibility: Can we utilize the object-aware attributes of DINO to learn robust spatio-temporal dependencies and thereby produce coherent object segmentations in videos?

In this paper, we delve into the potential of this idea with a simple architecture. 
Based on DINO pretrained ViT, we introduce a single spatio-temporal Transformer block to process frame-wise DINO features and calculate spatio-temporal attention maps. Remarkably, these attention maps in Fig.~\ref{fig:teaser}(c) showcase the ability to coarsely discriminate and track object parts even with random initialization. 
Based on this observation, we can roughly estimate the patches belonging to the same object over time.
Then, we encourage both semantic and motion alignment among patches belonging to the same object while distinguishing those from different objects in a fully self-supervised manner. 
The training significantly diminishes noise in the learned spatio-temporal dependencies as displayed in Fig.~\ref{fig:teaser}(d). 
In inference, we directly apply Hierarchical Clustering~\cite{johnson1967hierarchical} to these spatio-temporal attention maps to derive object segmentation masks within a sequence. Interestingly, this naive clustering method yields unexpectedly competitive results. Our method greatly outperforms slot attention-based methods on various datasets, e.g., \textbf{9} points improvement over SOLV~\cite{aydemir2023self} on DAVIS-17, \textbf{12} points improvement over SMTC~\cite{qian2023semantics} on YTVIS-19.
We argue that clustering on the whole spatio-temporal attention maps, instead of frame-by-frame feature clustering as in slot attention ensures the temporal coherence of produced segmentations and guarantees the object's permanence.

In summary, our proposed method \textbf{BA} (short for \textbf{B}etrayed by \textbf{A}ttention)  offers three distinct advantages: (1) BA solely requires RGB frames and operates entirely under self-supervision without the need for external modality. This characteristic renders our method highly accessible and adaptable in real-world scenarios; (2) The architecture is remarkably simple and systematically efficient. We only introduce a single learnable Transformer block and harness a parameter-free clustering algorithm to discover the objects. 
It facilitates easy generalization to multi-object scenes, without the constraints imposed by slot attention;
(3) Our method achieves state-of-the-art results across a bunch of unsupervised video multi-object segmentation benchmarks, such as MOVi-E~\cite{greff2021kubric}, DAVIS-17-Unsupervised~\cite{pont20172017}, and YouTube-VIS-19~\cite{Yang2019vis}. Notably, we surpass the previous state-of-the-art method, TimeT~\cite{salehi2023time}, by 6.6 points on YTVIS-19 and by 6.1 points on DAVIS-17 in terms of the FG-ARI metric. 

\section{Related Work}
\label{sec:rw}
\noindent\textbf{Video Object Segmentation} aims to segment objects coherently in a video sequence~\cite{caelles2017one, fan2019shifting, dutt2017fusionseg, guo2024animatediff, guo2023sparsectrl,  perazzi2017learning, hu2018videomatch, li2018video, bao2018cnn, johnander2019generative}. 
In video object segmentation (VOS), there are two prevalent protocols for evaluating the learned models: Semi-supervised VOS and Unsupervised VOS.
In semi-supervised VOS, the algorithm is provided with the object masks in the first frame, and required to track them in subsequent frames. In contrast, unsupervised VOS aims to identify and segment salient objects from the background without any specific reference.
In this paper, we focus on the more challenging unsupervised VOS setting without using any kind of manual annotations in either training or inference.
Recently, there emerges a line of self-supervised algorithms for unsupervised VOS~\cite{xie2019object, Yang_2019_CVPR, NEURIPS2019_140f6969, wang2019unsupervised, Yang_2021_ICCV, Yang_2021_CVPR, choudhury2022guess, kipf2022conditional, ye2022deformable}.  
CIS~\cite{Yang_2019_CVPR} facilitates fully unsupervised motion segmentation, discarding object mask supervision during training. By formulating a min-max game of mutual information, the generator is motivated to create segmentations that effectively distinguish foreground objects from the background.
AMD~\cite{liu2021emergence} minimizes the warping synthesis error to train appearance and motion pathways without any supervision.
GWM~\cite{choudhury2022guess} uses a single RGB image as input with optical flow acting as supervision to highlight moving areas.
MG~\cite{Yang_2021_ICCV} and OCLR~\cite{xie2022segmenting} solely leverage optical flow as input to generate object-centric layered representations, with each layer indicating a potential object. 
Despite the impressive performance of state-of-the-art self-supervised VOS methods~\cite{qian2023semantics, salehi2023time, aydemir2023self} in discovering single objects, they still face limitations when it comes to solving multi-object discovery tasks.
In contrast, by leveraging the rich object cues in attention maps, our method can discern multiple objects in real-world scenarios.

\noindent\textbf{Self-supervised Spatio-temporal Correspondence learning} usually leverages free temporal supervision signals in videos to learn representations that facilitate accurate pixel or object tracking across space and time~\cite{lai2020mast, Lai19,vondrick2018tracking,wang2019learning,jabri2020space,bian2022learning, hu2022semantic,xu2021rethinking,caron2021emerging, li2019joint, ding2022motion}. 
Vondrick et al.~\cite{vondrick2018tracking} employ the natural temporal color coherence to train a colorization model on grayscale videos, thereby establishing fine-grained correspondence between current and future frames.
CRW~\cite{jabri2020space} presents a self-supervised learning approach for dense correspondence in raw videos. It uses space-time graph-based random walks and cycle consistency to implicitly supervise chains of comparisons. 
Hu et al.~\cite{hu2022semantic} independently learn semantics and temporal correspondence from two pathways and fuse them at a later stage. UME~\cite{li2023unified} respectively designs short-term appearance and long-term semantic consistency to learn generalizable correspondence.
Taking a step further by integrating high-level semantics with low-level temporal correlation, SMTC~\cite{qian2023semantics} develops two-stage slot attention to establish dense correspondence with more emphasis on foreground objects.
In this work, we explicitly initialize the spatio-temporal correspondence on top of the DINO pretrained Transformer. This initialization effectively guides the learning process with a self-supervised pixel-level consistency.

\noindent\textbf{Object-aware DINO Features} have shown efficacy in object localization tasks, particularly in image domain~\cite{simeoni2021localizing, bielski2022move, hamilton2022unsupervised, ziegler2022self, wang2024barleria, melas2022deep, zadaianchuk2023unsupervised, Wang_2023_CVPR}. 
DINO's original paper~\cite{caron2021emerging} reveals that the learned representations and attention maps carry substantial object cues that can be harnessed for object discovery.
The groundbreaking work, LOST~\cite{simeoni2021localizing}, utilizes these features from DINO to perform object segmentation by constructing a graph wherein objects are segmented using the inverse degrees of nodes. 
TokenCut~\cite{wang2022tokencut2} employs DINO features for applying the Normalized Cut algorithm~\cite{shi2000normalized}, thereby obtaining foreground segments in an image.
Furthermore, CutLER~\cite{Wang_2023_CVPR} and MOST~\cite{rambhatla2023most} can localize multiple objects without any supervision based on the DINO features.
Motivated by the successful application of DINO features in image settings, recent works have expanded their use to video segmentation tasks~\cite{wang2022tokencut2, qian2023semantics, aydemir2023self, salehi2023time,wang2024videocutler}. SMTC~\cite{qian2023semantics} utilizes the rich semantics and correspondence cues in the DINO features to provide a reliable reference for object decomposition in consecutive frames. TimeT~\cite{salehi2023time} introduces a feature forwarding process to propagate dense DINO features across time and establish temporal consistency. 
Most of the existing works focus on leveraging DINO feature vectors for optimization and adaptation to videos. While in our work, we lay more emphasis on the self-attention maps. We extend the frame-wise DINO attention maps to spatio-temporal maps, which explicitly encode space-time structural dependencies and help coherent object discovery in videos.

\section{Method}

\subsection{Preliminary on DINO Attention Maps}
\label{pre}
The self-supervised ViT, as achieved by DINO~\cite{caron2021emerging}, has exhibited a rich emergence of objectness in the attention maps. To elucidate this, we visualize the self-attention map from the final Transformer block of the DINO pretrained ViT-S/8 as a representative. As illustrated in Fig.~\ref{dino}(b), an image patch exhibits a high degree of attention dependencies with patches of the same object. Conversely, image patches of distinct objects tend to present different attention distributions. 
Encouraged by this observation, we pose the following question: Could it be possible to directly harness these attention maps to achieve object segmentation without relying on annotations?

\begin{table}[]
\begin{minipage}{0.55\linewidth}
    \centering
    \small
    \begin{tabular}{cc p{0.8cm}<{\centering} p{0.8cm}<{\centering} p{0.8cm}<{\centering}}
    \toprule
    \multirow{2}{*}{\shortstack{Clustering\\ Metric}} & Cosine similarity & \multicolumn{3}{c}{KL divergence} \\\cmidrule(r){2-2}\cmidrule(r){3-5}
    & $F$ & $A$ & $A_v$ & $\Tilde{A}_v$ \\
    \midrule
    DAVIS-16 & 47.4 & 60.3 &  51.1 & 75.4 \\
    DAVIS-17 & 14.7 &  26.7 & 20.9 & 39.2 \\
    \bottomrule
    \end{tabular}
    \caption{Preliminary video object segmentation results using varying clustering metrics. $F$ represents features pretrained with DINO. $A$ denotes per-frame attention maps from DINO, while $A_v$ and $\Tilde{A}_v$ denote spatio-temporal attention maps produced by a randomly initialized and a final learned temporal correlator respectively, on top of DINO. 
    The IoU ($\mathcal{J}$ score) is reported for both DAVIS-16 and DAVIS-17-Unsupervised datasets.}
    \label{dino}
\end{minipage}
\hspace{0.02\linewidth}
\begin{minipage}{0.42\linewidth}
    \includegraphics[width=\linewidth]{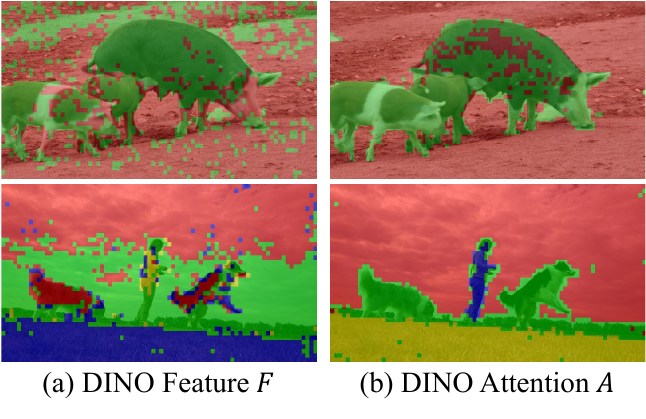}
    \captionof{figure}{Visualizations of the clustering results. The left column is the results of DINO features $F$, the right column is the results of DINO attention $A$. $F$ results in much noisier clusters, while $A$ distinguishes different classes of objects.}
    \label{dionvis}
\end{minipage}
\end{table}

In response to this question, we conduct a straightforward initial exploration with these attention maps.  
We denote the attention map of the last Transformer block as $A\in\mathbb{R}^{HW\times HW}$ and its output features $F\in\mathbb{R}^{HW\times C}$, where $H$, $W$ and $C$ respectively denote the height, width, and channel dimension of the feature map\footnote{Here we omit the \texttt{CLS} token, and average the multi-head attention along the head dimension.}. 
For an intuitive comparison, we apply unsupervised clustering on both attention maps $A$ and features $F$ to yield the object segmentation masks. 
For the detailed procedure, please refer to the Sec.~\ref{ours_vos}.
Then, we report frame-wise evaluation results on the video object segmentation datasets DAVIS-16~\cite{perazzi2016benchmark} and DAVIS-17-Unsupervised~\cite{pont20172017} in Table~\ref{dino}. 
Interestingly, we observe that the attention maps significantly excel in decoupling object components compared to features, exhibiting an improvement of over \textbf{10} points on both datasets in terms of IoU.  Surprisingly, a spatio-temporal attention map $A_v$ generated by a randomly initialized Transformer block on top of DINO also outperforms the frame-wise DINO feature $F$. And it is further demonstrated in Fig.~\ref{dionvis}. The DINO features lead to very noisy clusters, while attention maps can discriminate objects of different categories. We speculate this is because the attention maps preserve intact correspondences, which are more suitable for fine-grained tasks such as object grouping and discrimination. Therefore, creating spatio-temporal structural dependencies that expand beyond spatial DINO attention maps appears promising for video object segmentation tasks.
\begin{figure*}[t]
    \centering
    \includegraphics[width=\linewidth]{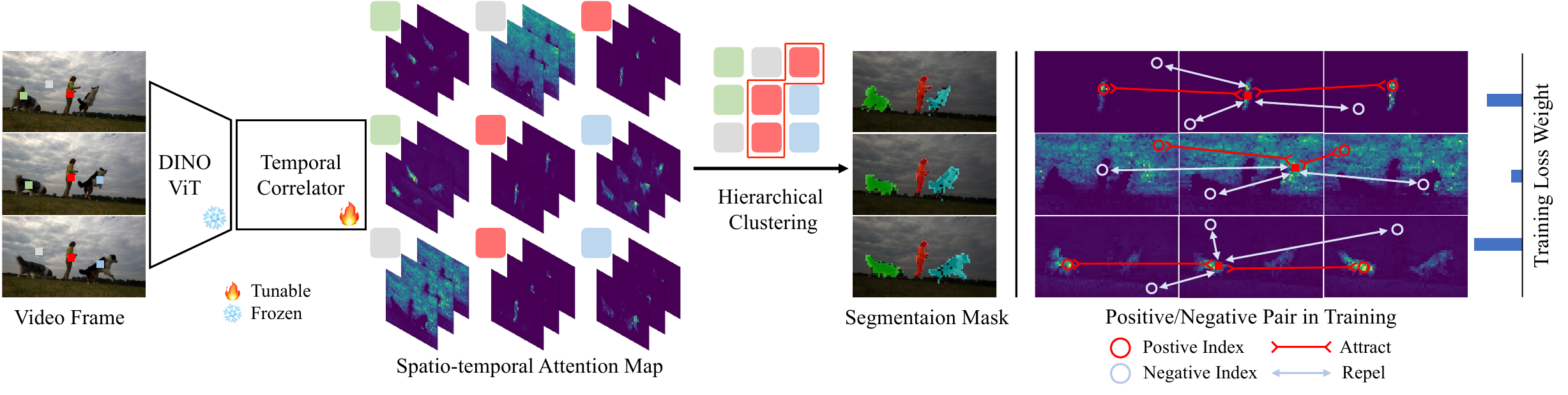}
    \caption{Our architecture BA overview. Given the video frames and a DINO pretrained Transformer, we first use a temporal correlator to construct the spatio-temporal correspondence. We then utilize these attention maps as a clustering metric and apply hierarchical clustering across all frames to generate segmentation masks. During training, for each patch, we sample a positive/negative set and assign an importance weight based on its corresponding attention map. We promote alignment within the positive set while differentiating the representations from the negative set. The final loss, normalized with importance weights, only trains the temporal correlator with the DINO ViT remaining frozen.}
    \label{fig:arch}
\end{figure*}

\subsection{Our Simple Architecture for VOS}
\label{ours_vos}
Motivated by the frame-wise preliminary result, we introduce a temporal correlator instantiated with a spatio-temporal Transformer block~\cite{bertasius2021space} on top of the DINO pretrained ViT backbone, to extend the self-attention calculation to the spatio-temporal domain. 
Specifically, given the frame-wise DINO features $\{F_1,\dots,F_T\}$ of $T$ frames, we process them through a spatio-temporal Transformer block $\Phi_{\text{st}}$. It consists of one standard ViT encoder layer, with self-attention computed over the spatio-temporal grids. Specifically, we concatenate the frame-wise DINO features along the temporal dimension with the addition of learnable temporal positional embeddings $\texttt{pe}$, i.e., \(F = \{F_1 + \texttt{pe}_1, \ldots, F_T + \texttt{pe}_T\} \in \mathbb{R}^{THW\times C}\) and then linearly project $F$ to obtain the queries \(Q=f_Q(F)\), keys \(K= f_K(F)\), and values \(V=f_V(F)\).
Then, we can obtain the self-attention map $A_v\in\mathbb{R}^{THW\times THW}$ and temporally fused features $F_v\in \mathbb{R}^{THW\times C}$: 
\begin{equation}
A_v = \text{softmax}\left(\frac{Q K^{\top}}{\sqrt{C}}\right), \quad
F_v = A_v V=\{\Tilde{F}_1, \ldots, \Tilde{F}_T\}.
\end{equation}
In this way, we establish spatio-temporal structural dependencies beyond frame-wise DINO features and have the potential to improve object permanence across the entire time span.

Since $A_v[i]\in\mathbb{R}^{THW}$ is a proximity distribution that describes the structural dependencies between the $i$-th patch and all other positions, we adopt symmetric KL divergence to measure the distance $\mathcal{M}\in \mathbb{R}^{THW\times THW}$ between any two patches as:
\begin{align}
    \mathcal{M}[i,j] = \mathcal{D}_{KL}(A_v[i]||A[j]) + \mathcal{D}_{KL}(A_v[j]||A_v[i]),
    \label{kldiv}
\end{align}
where $\mathcal{D}_{KL}$ denotes KL divergence. 
Based on this distance matrix $\mathcal{M}$, we then apply Hierarchical Clustering~\cite{johnson1967hierarchical} to obtain the object segmentation masks. Please consult our appendix for a detailed implementation of the hierarchical clustering algorithm.
In this way, we iteratively merge $THW$ patches into $N$ centroids $A_c\in\mathbb{R}^{N\times THW}$. 
Since each centroid likely represents the spatial distribution of an object, we consider the cluster assignments as potential candidates for the object segmentation mask.
Specifically, we calculate the KL divergence distance between the $A_v$ and $A_c$ as $D\in\mathbb{R}^{THW\times N}$, and use \texttt{argmin} operation to produce the cluster assignments $Z$ within the video sequence:
\begin{align}
    Z = \texttt{argmin}(D,\texttt{dim=1})\in\{1,...,N\}^{THW}.
\end{align}
Then, we employ a bipartite matching mechanism, following the standard evaluation protocol in DAVIS-17 Challenge~\cite{pont20172017}, to match the predicted masks with ground truth for evaluation.

A recent study, DiffSeg~\cite{tian2023diffuse}, used hierarchical clustering to process attention maps from a stable diffusion model, but it operated solely in the image domain. Our work extends the powerful image foundation model to the spatio-temporal domain, demonstrating that accurate spatio-temporal correspondence can be learned without supervision. Additionally, our method does not rely on hierarchical clustering. As shown in Table~\ref{tab:cluster_metric}, using K-means clustering also achieves competitive performance, indicating that the learned spatio-temporal correspondence is effective across various clustering strategies.

Besides, there are two noteworthy aspects in our clustering stage. First, in the clustering process, we only set a distance threshold hyper-parameter without specifying the number of cluster centroids and the hierarchical clustering algorithm can adaptively generate different numbers of clusters according to the complexity of the video. Second, our reference to the entire spatio-temporal distribution enables coherent object segmentation across multiple frames. As a result, the generated segmentation masks effortlessly track corresponding objects over time without extra association procedures.

\subsection{Training}
\label{train}
Noticing that even a randomly initialized attention map $A_v$ shown in Figure~\ref{dino}(c), exhibits an initial ability to differentiate parts and track objects over time, we leverage these maps as indicators for positive and negative sets sampling. 
Mathematically, given the attention maps $A_v\in\mathbb{R}^{THW\times THW}$, we first \texttt{reshape} $A_v$ into shape $THW\times T\times HW$ then use \texttt{top-k} operation to retrieve $K_p$ position indexes in each frame with highest attention scores:
\begin{align}
    I_p = \texttt{top-k}(\texttt{reshape}(A_v), K_p, \texttt{dim=-1}).
    \label{index}
\end{align}
In this way, we obtain a position index set $I_p\in\mathbb{R}^{THW\times T\times K_p}$, which represents positions potentially associated with the same object as the $i$-th patch. The \texttt{reshape} operation ensures that we consistently extract the same number of patches in each frame. This, in turn, guarantees effective object tracking within consecutive frames and promotes consistent temporal correspondence learning. 
Similarly, we perform this operation to find $K_n$ positions in each frame with the lowest attention scores $I_n\in\mathbb{R}^{THW\times T\times K_n}$. Each $I_n[i]$ represents positions least likely to belong to the same object as the $i$-th patch.
After obtaining the positive and negative index sets, we apply self-supervised correspondence learning in two aspects: semantic consistency and dynamic motion consistency.

\noindent\textbf{Semantic consistency.}
Intutively, we take the temporally fused features $F_v=\{\Tilde{F}_1,...,\Tilde{F}_T\}\in\mathbb{R}^{THW\times C}$ as the semantic representation. 
We promote the semantic alignment between the features of the same object and distinguish features from different objects. 
Given the $i$-th patch, we first refer to the positive pair indexes $I_p[i]$ in Eq.~\ref{index} to gather the features that are likely from the same object. Then we calculate the cosine distance between the query feature and all positive feature vectors, and produce the distance matrix $\mathcal{S}_p[i]\in\mathbb{R}^{TK_p}$. 
Similarly, we gather distinct features using $I_n[i]$, and form negative distance matrix $\mathcal{S}_n[i]$. We employ a simple margin loss to enforce higher consistency between corresponding object areas:
\begin{align}
    \mathcal{L}_{\text{s}}[i] = \sum_{j=1}^{TK_p}\sum_{k=1}^{TK_n}\max(\mathcal{S}_p[i,j]-\mathcal{S}_n[i,k]+\lambda_1, 0),
\end{align}
where $\lambda_1$ is the margin hyper-parameter and we set $\lambda_1=0.8$ as default.

\noindent\textbf{Dynamic motion consistency.}
Regarding dynamic motions, rather than calculating optical flow, inter-frame feature correlations, or latent cost volumes as motion representations, we turn to highly accessible attention maps $A_v$. These maps exhibit spatio-temporal correlations of specific patches, serving as an effective latent representation of temporal dynamics. 
Therefore, it is feasible to directly employ these attention maps to bolster motion consistency between corresponding objects.
Similar to the semantic consistency learning, given the attention map of $i$-th patch $A_v[i]$ as a query, we also leverage the positive (negative) pair indexes $I_p[i]$ ($I_n[i]$) to gather the attention maps of the corresponding (distinct) objects. Then we calculate the symmetric KL divergence as in Eq.~\ref{kldiv} between the query and all positive (negative) attention maps, obtaining the distance matrix $\mathcal{M}_p[i]\in\mathbb{R}^{TK_p}$ and $\mathcal{M}_n[i]\in\mathbb{R}^{TK_n}$. We utilize the same margin loss for optimization, with $\lambda_2=1.0$ as the margin hyper-parameter:
\begin{align}
    \mathcal{L}_{\text{m}}[i] = \sum_{j=1}^{TK_p}\sum_{k=1}^{TK_n}\max(\mathcal{M}_p[i,j]-\mathcal{M}_n[i,k]+\lambda_2, 0).
\end{align}

\noindent\textbf{Overall objectives.}
Besides, we evaluate the importance of each spatio-temporal patch and lay more emphasis on the informative regions. Formally, we employ the entropy of the attention maps as a measure of the information content within each patch:
\begin{align}
    e[i] = \sum_{k=1}^{THW} -A_v[i,k]\log A_v[i,k],
\end{align}
with $e[i]$ denoting the entropy of $i$-th patch. A higher entropy value indicates more ambiguous spatio-temporal correlations, which provide less information. Hence, we perform \texttt{softmax} normalization on the reversed entropy to produce the importance weight $w$ for each patch as:
\begin{align}
    w[i] = \frac{\exp(-e[i])}{\sum_{j=1}^{THW}\exp(-e[j])}.
\end{align}
Then, the overall learning objectives on semantic and motion consistency can be formulated as the weighted summation over all spatio-temporal patches:
\begin{align}
    \mathcal{L} = \sum_{i=1}^{THW}w[i](\mathcal{L}_{\text{s}}[i]+\mathcal{L}_{\text{m}}[i]).
\end{align}
Since we only introduce one additional temporal correlator and an entirely parameter-free clustering strategy, we exclusively tune a single standard spatio-temporal Transformer block with 1.6M parameters and freeze the DINO encoder throughout the training process. 
This straightforward method not only maintains a minimal number of learnable parameters but also yields remarkable results in video object segmentation.

\section{Experiments}
\subsection{Datasets}
For training, we adopt the challenging video dataset YouTube-VIS-19~\cite{Yang2019vis} for video instance segmentation.
YouTube-VIS-19 consists of 2,883 high-resolution YouTube videos and each video usually contains multiple object instances of distinct semantics. 
During the inference, we evaluate our method on unsupervised multiple object segmentation. We use the synthetic MOVi-E~\cite{greff2021kubric} dataset and real-world datasets like DAVIS-17-Unsupervised~\cite{pont20172017} and YouTube-VIS-19~\cite{Yang2019vis}, reporting Foreground Adjusted Rand Index (FG-ARI) and mIoU. For MOVi-E, we integrate its training set into our training due to significant distribution shifts. For the rest, we maintain solely training on the YouTube-VIS-19 set.
Note that we do not present any mask annotation to the model during either the training or inference stages.

\subsection{Implementation Details}
\label{imp}
In training, we sample $T=3$ frames with temporal stride $4$ as the input clip. Each frame is augmented with random crop and color jitter and resized to $192 \times 384$. 
We adopt DINO pretrained ViT-S/8~\cite{caron2021emerging} as the frame encoder, which is then followed by a single spatio-temporal Transformer Encoder block with 8 heads as temporal correlator $\Phi_{\text{st}}$. We set the number of sampled positive and negative pairs to $K_p=10$ and $K_n=50$ in default. For optimization, we adopt AdamW~\cite{loshchilov2018decoupled} with a learning rate $1\times 10^{-4}$, and train the model for a total of 30k iterations with a batch size of 16.
In the inference stage, our model can be flexibly applied to video sequences of arbitrary lengths. For inference, we set the distance threshold in hierarchical clustering to 1.0. We follow the standard evaluation protocol outlined in the DAVIS-17 challenge~\cite{pont20172017} to match the predicted masks with the ground truth.


\begin{table*}[!h]
    \centering
    \caption{Quantitative results on multiple object video segmentation. For MOVi-E and YouTube-VIS-19 datasets, we report FG-ARI and mIOU. Besides, we report Region Similarity $\mathcal{J}$ and Contour Accuracy $\mathcal{F}$ on DAVIS-17-Unsupervised. $\star$ The original paper of TimeT~\cite{salehi2023time} includes the IoU of the background category, we rerun the evaluation to exclude background here. $\dagger$ We use DINOv2 ViT-S/14 as our frozen frame encoder.} 
    \begin{tabular}{l p{1.2cm}<{\centering} p{1.07cm}<{\centering} p{1.2cm}<{\centering} p{1.07cm}<{\centering} p{1.2cm}<{\centering} p{1.07cm}<{\centering} p{1.07cm}<{\centering} p{1.07cm}<{\centering}}
    \toprule
        {} & \multicolumn{2}{c}{MOVi-E} & \multicolumn{2}{c}{YTVIS-19} & \multicolumn{4}{c}{DAVIS-17}  \\
        \cmidrule(r){2-3}
        \cmidrule(r){4-5}
        \cmidrule(r){6-9}
        Model  & FG-ARI & mIoU & FG-ARI & mIoU &  FG-ARI &  $\mathcal{J}\&\mathcal{F}$ & $\mathcal{J}$ & $\mathcal{F}$  \\ \midrule
        SAVi~\cite{kipf2022conditional} & 42.8 & 16.0 & 11.1 & 12.7 & - & - & - & - \\
        STEVE~\cite{singh2022simple} & 50.6 & 26.6 & 20.0 & 20.9 & - & - & - & - \\
        OCLR~\cite{xie2022segmenting} & - & - & 15.9 & 32.5 & 14.7 & - & 34.6 & -  \\
        VideoSAUR~\cite{zadaianchuk2023objectcentric} & 73.9 & 35.6 & 39.5 & 29.1 & - & - & - & - \\
        SOLV~\cite{aydemir2023self} & 80.8 & - & 29.1 & 45.3 & 32.2 & - & 30.2 & -  \\
        SMTC~\cite{qian2023semantics} & - & - & 31.4 & 38.8 & 33.3 & 40.5 & 36.4 & 44.6 \\
        TimeT$^\star$~\cite{salehi2023time} & - & - & 37.9 & 40.4 & 35.5 & 40.0 & 35.8 & 44.2 \\\midrule
        BA (ours) & 83.4 & 40.2 & 44.3 & 50.1 & 40.1 & 43.9 & 39.2 & 48.6  \\
        BA$^\dagger$ (ours) & 84.4 & 40.7 & 44.5 & 50.1 &  41.6 & 43.7 & 39.4 & 48.0  \\
        \bottomrule
    \end{tabular}
    \label{tab:multiple}
\end{table*}

\subsection{Comparison with State-of-the-art}

We evaluate our method on challenging multi-object segmentation, which aims to segment different objects in a temporally coherent manner. Among the compared methods, SAVi~\cite{kipf2022conditional} and OCLR~\cite{xie2022segmenting} employ optical flow as supervision to discriminate motions of different objects. 
STEVE~\cite{singh2022simple}, VideoSAUR~\cite{zadaianchuk2023objectcentric}, SOLV~\cite{aydemir2023self} and SMTC~\cite{qian2023semantics} develop variants of slot attention with temporal constraints to bind to distinct instances. TimeT~\cite{salehi2023time} propagates dense DINO features across time to enhance temporal consistency and applies clustering on learned features to produce object segmentations. In contrast, our method directly uses spatio-temporal attention maps for object segmentation. As shown in Table~\ref{tab:multiple}, our method achieves state-of-the-art results on both synthetic (MOVi-E) and realistic video datasets (YTVIS-19 and DAVIS-17), indicating promising generalization ability. 
Specifically, comparing to the prior arts, we gain an approximately 5-point advantage over both  VideoSAUR~\cite{zadaianchuk2023objectcentric} on YTVIS-19 and SMTC~\cite{qian2023semantics} and TimeT~\cite{salehi2023time} on DAVIS-17 in terms of FG-ARI. 
The superior results under our simple architecture reveal the feasibility of utilizing the spatio-temporal dependencies to mine object cues and derive reliable video object segmentation results. 
Besides, we also report the results using DINOv2 ViT-S/14 as the frame encoder. Despite the larger patch size, our method still achieves competitive performance. This demonstrates the robustness of our approach to pretraining models.

\begin{figure}
    \centering
    \includegraphics[width=\linewidth]{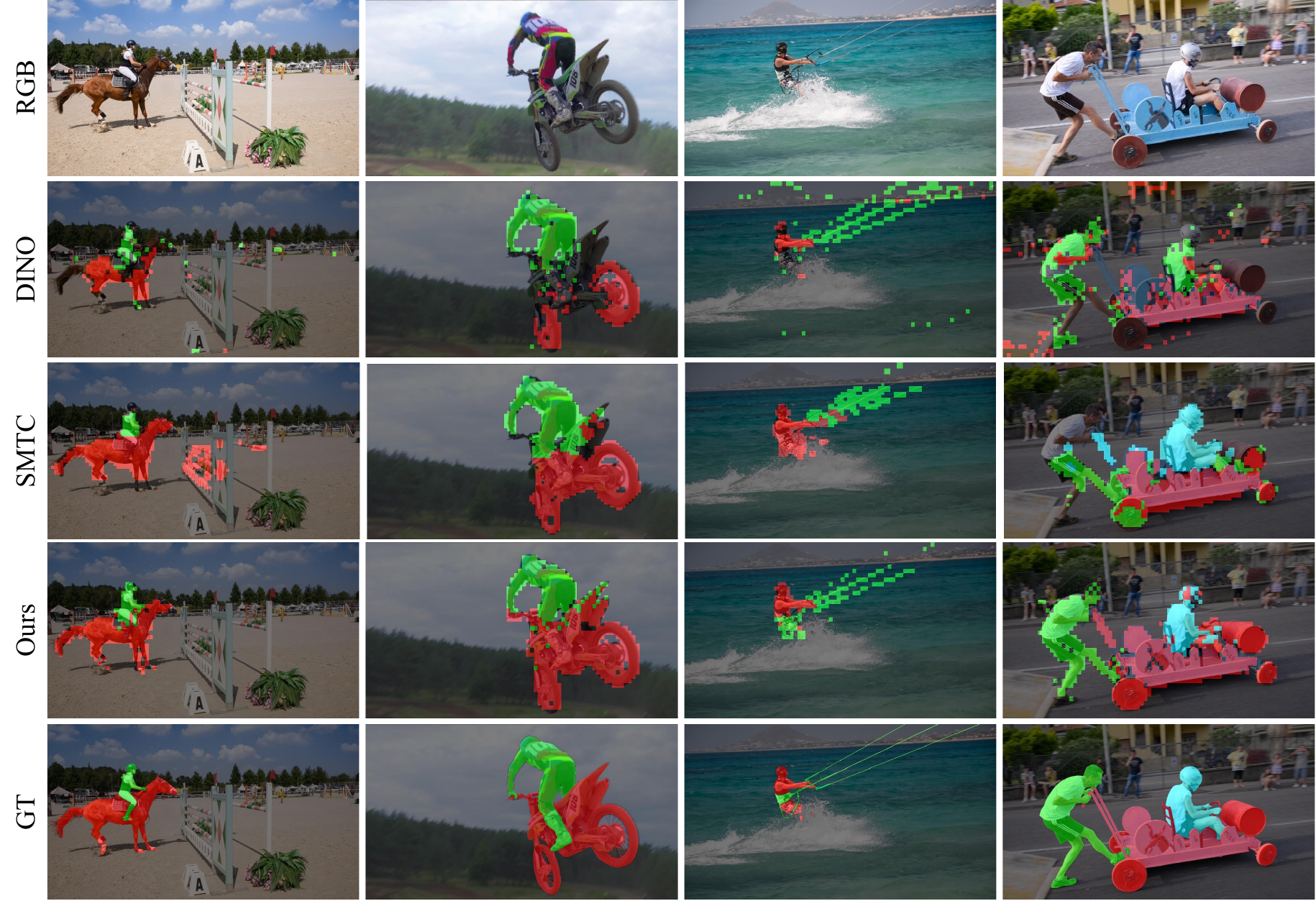}
    \caption{Qualitative comparison with DINO~\cite{caron2021emerging} and SMTC~\cite{qian2023semantics} on DAVIS-17-Unsupervised. Different color denotes different clusters.}
    \label{fig:compare}
\end{figure}

\noindent\textbf{Qualitative Comparison.}
To visually demonstrate the effectiveness of our proposed method, we contrast our segmentation results with those of the DINO counterpart and prior arts SMTC~\cite{qian2023semantics} in Fig.~\ref{fig:compare}. For the DINO baseline, we cluster the patches through the attention map $A$. As can be observed, our method is able to segment with significantly less noise. For instance, in the motocross-jump sequence, our method nearly outlines the entire motorbike and the rider, whereas DINO misses most of the bike's body and only identifies its tyre. 
This shows that inferring the masks by associating a collection of RGB features and constructing spatio-temporal correspondences greatly enhances the segmentation result.
Additionally, with the benefit of temporal correspondence, our model can differentiate between distinct object instances with the same semantics. Consider the soapbox sequence displayed in the right column; our method distinguishes two different people - one running behind the soapbox and another sitting in the soapbox throughout the video. This level of distinction cannot be obtained from DINO's spatial attention map, further emphasizing the effectiveness of introducing rich spatio-temporal relations in video segmentation tasks.

\subsection{Ablation Study}
In the ablation study, we utilize DINO ViT-S/8 as the frozen backbone. We report FG-ARI and IoU on YouTube-VIS-2019 and $\mathcal{J}\&\mathcal{F}$ on DAVIS-17-Unsupervised for multi-object segmentation. 

\noindent\textbf{Training objective $\mathcal{L}_{\text{s}}$, $\mathcal{L}_{\text{m}}$, and $w$.} 
To understand the contribution of each component to the VOS performance, we individually evaluate each loss item as depicted in Table~\ref{tab:loss}. 
Compared to the baseline without training, the inclusion of semantic consistency loss $\mathcal{L}_{\text{s}}$ with entropy-based weighting mechanism $w$ (second row) elevates FG-ARI from 23.4 to 30.1 on YouTube-VIS-19. Furthermore, the motion consistency loss $\mathcal{L}_{\text{m}}$ (third row) yields even more substantial improvements, resulting in FG-ARI reaching 39.9. This aligns with our preliminary observations that dynamic motions play a more significant role than semantics in fine-grained object segmentation.
Combining all components (fourth row) provides the best results, highlighting the synergy between the components.
Additionally, removing the weighting function $w$ (fifth row) slightly decreases the performance on both two benchmarks, affirming the value of entropy normalization to prioritize informative regions.

\noindent\textbf{Clustering metrics and strategy.} To further explore whether our clustering strategy optimally leverages the learned spatio-temporal properties, we select four types of clustering metrics for comparison. Specifically, we adopt both feature vectors and attention maps directly from DINO and those from our learned spatio-temporal Transformer block. Table~\ref{tab:cluster_metric} reveals three key observations. First, our learning paradigm consistently outperforms the off-the-shelf DINO in terms of both features and attention maps, improving all metrics by over 10 points. This strongly supports our motivation that establishing spatio-temporal correspondence is crucial for video object segmentation. Second, utilizing the intermediate attention maps directly for clustering proves to be more effective than employing the final features. We speculate this occurs because the attention maps encode fine-grained structural dependencies between various positions, whereas feature vectors primarily capture high-level semantics. Third, the hierarchical clustering strategy is not mandatory for success. We also achieve similar performance by replacing it with K-means clustering, where we set the number of centroids to 8, indicating that BA distills general object-aware representations across time.

\begin{table}[!h]
\begin{minipage}{0.45\linewidth}
    \centering
    \caption{Ablation on different combinations of semantic consistency loss $\mathcal{L}_{\text{s}}$, motion consistency loss $\mathcal{L}_{\text{m}}$ and entropy weight $w$. The first row represents the results obtained from a randomly initialized spatio-temporal Transformer block.} 
    \begin{tabular}{ccccccc}
    \toprule
        {} & {} & {} & \multicolumn{2}{c}{YTVIS-19} & \multicolumn{2}{c}{DAVIS-17} \\
        \cmidrule(r){4-5}
        \cmidrule(r){6-7}
        $\mathcal{L}_{\text{s}}$ & $\mathcal{L}_{\text{m}}$ & $w$ & FG-ARI & mIoU & $\mathcal{J}\&\mathcal{F}$ & FG-ARI \\
        \midrule
        - & - & - & 23.4 & 25.5 & 23.1 & 18.4 \\
        \checkmark & \xmark & \checkmark & 30.1 & 32.4 & 29.6 & 27.5 \\
        \xmark & \checkmark & \checkmark & 39.9 & 46.1 & 38.6 & 37.3 \\
        \checkmark & \checkmark & \checkmark & 44.3 & 50.1 & 43.9 & 40.1 \\
        \checkmark & \checkmark & \xmark & 43.1 & 48.1 & 42.5 & 38.1 \\
        \bottomrule
    \end{tabular}
    \label{tab:loss}
\end{minipage}
\hspace{0.02\linewidth}
\begin{minipage}{0.5\linewidth}
    \centering
    \small
    \caption{Ablation on different clustering metrics. Features $F$ and attention maps $A$ are directly obtained from DINO. $F_v$ and $A_v$ are generated from our learned spatio-temporal Transformer block. $A_v$ and $A_v^*$ respectively denote the hierarchical clustering and K-means clustering results.}
    \begin{tabular}{ccccc}
    \toprule
        \multirow{2}{*}{\shortstack{Clustering\\ Metric}} & \multicolumn{2}{c}{YTVIS-19} & \multicolumn{2}{c}{DAVIS-17} \\
        \cmidrule(r){2-3}
        \cmidrule(r){4-5}
         & FG-ARI & mIoU & $\mathcal{J}\&\mathcal{F}$ & FG-ARI \\
        \midrule
        $F$ & 22.1 & 23.8 & 16.1 & 19.4 \\
        $F_v$ & 33.9 & 31.4 & 33.4 & 32.1 \\
        $A$ & 30.1 & 34.8 & 29.3 & 30.5 \\
        $A_v$ & 44.3 & 50.1 & 43.9 & 40.1 \\
        $A_v^*$ & 42.5 & 48.5 & 41.3 & 39.3 \\
    \bottomrule
    \end{tabular}
    \label{tab:cluster_metric}
\end{minipage}
\end{table}

\begin{table}[t]
\begin{minipage}{0.45\linewidth}
    \centering
    \small
    \caption{Ablation on the number of frames $T$ and temporal stride. When sampling more than 3 frames with a larger stride, the method exhibits a diminishing effect.}
    \begin{tabular}{cccccc}
    \toprule
        {} & {} & \multicolumn{2}{c}{YTVIS-19} & \multicolumn{2}{c}{DAVIS-17} \\
        \cmidrule(r){3-4}
        \cmidrule(r){5-6}
        $T$ & Stride & FG-ARI & mIoU & $\mathcal{J}\&\mathcal{F}$ & FG-ARI \\
        \midrule
        3 & 1 & 37.4 & 43.6 & 37.1 & 33.6 \\
        3 & 4 & 44.3 & 50.1 & 43.9 & 40.1 \\
        3 & 8 & 44.2 & 50.2 & 43.3 & 40.7 \\
        5 & 4 & 44.8 & 50.6 & 44.3 & 41.5 \\
        \bottomrule
    \end{tabular}
    \label{tab:frame}
\end{minipage}
\hspace{0.02\linewidth}
\begin{minipage}{0.5\linewidth}
    \centering
    \small
    \caption{Ablation on different motion representations in calculating motion consistency loss $\mathcal{L}_{\text{m}}$. We compare using the self-attention matrix, simple local and global feature correlation as the latent motion representation.}
    \begin{tabular}{lcccc}
    \toprule
        {} & \multicolumn{2}{c}{YTVIS-19} & \multicolumn{2}{c}{DAVIS-17} \\
        \cmidrule(r){2-3}
        \cmidrule(r){4-5}
        MotionRep & FG-ARI & mIoU & $\mathcal{J}\&\mathcal{F}$ & FG-ARI \\
        \midrule
        Local Corr. & 30.5 & 33.8 & 29.6 & 27.7 \\
        Global Corr. & 39.1 & 49.2 & 36.4 & 33.9 \\
        Self-Attention & 44.3 & 50.1 & 43.9 & 40.1 \\
    \bottomrule
    \end{tabular}
    \label{motion}
\end{minipage}
\end{table}

\noindent\textbf{Number of frames $T$ and temporal stride.}
To investigate whether a wider temporal receptive field can reinforce temporal coherence and enhance object segmentations, we experiment by varying the number of input frames and temporal stride in training. Our results, displayed in Table~\ref{tab:frame}, reveal that the performance significantly deteriorates when the stride is very small, such as a temporal stride of 1 in the first row. However, it begins to plateau upon reaching 4. This is because a small temporal stride leads to very subtle temporal dynamics in consecutive frames, resulting in a trivial learning task while a larger temporal stride provides richer dynamics, effectively encouraging the model to capture temporal consistency and maintain object permanence.
Moreover, adding more frames results in diminishing returns, indicating that a modest temporal receptive field is adequate for sufficient temporal context in video object segmentation.

\noindent\textbf{Motion representations.}
In training, we use the spatio-temporal attention maps as latent motion representations for dynamic motion alignment $\mathcal{L}_{\text{m}}$. There are some alternatives to formulate this motion representation. Here we explore two variants in Table~\ref{motion}. The one is simple global feature correlation, which directly calculates the dot product between spatio-temporal feature vectors, i.e., $C_g=\Tilde{F}\Tilde{F}^{\top}\in\mathbb{R}^{THW\times THW}$. The other is local feature correlation, where we sample a sliding local window with size $T_p\times H_p\times W_p$ for each grid, and obtain the local correlation matrix $C_l\in\mathbb{R}^{THW\times T_pH_pW_p}$. It is obvious that using the self-attention maps as motion representation substantially outperforms the performance of the other two methods. Compared to the global correlation, the learnable parameters in attention layers enhance the modeling capacity, which enables the self-attention maps to capture more comprehensive spatio-temporal dependencies and serve as a better motion representation. The local correlation only captures temporal dynamics in limited reception fields, restricting the ability for long-term perception.


\section{Conclusion}
In this paper, we introduce a simple yet powerful approach for self-supervised video object segmentation. Specifically, our proposed architecture introduces only a single spatio-temporal Transformer block that ingests frame-wise DINO features and constructs the spatio-temporal correspondence through the attention layer. During training, we adopt a dual self-supervised consistency objective that encompasses both semantic consistency and dynamic motion consistency, supplemented with entropy normalization. In inference, we utilize hierarchical clustering on spatio-temporal attention maps to generate temporally coherent object segmentation masks. This extremely straightforward approach nevertheless delivers state-of-the-art results on challenging real-world multi-object segmentation tasks such as DAVIS-17-Unsupervised and YouTube-VIS-19.

\noindent\textbf{Limitation.} While promising, our method still has room for improvement. Firstly, the current spatio-temporal attention block cannot accommodate very long window-size frames due to the quadratic memory cost. 
Potentially, adopting memory mechanisms~\cite{zhou2024rmem,qian2024streaming} that are capable of handling long-duration data may prove more suitable. 
Secondly, neither pixel-level constraints nor spatio-temporal dependencies consider the granularity in the video. Incorporating multi-scale features~\cite{bian2022learning} may present a promising direction.

\noindent\textbf{Social Negative Impact.} Although our method produces coherent video object segmentation without annotation, it could also enhance deepfake video technologies, making it easier to produce false videos for malicious intent. It is important to regulate the use of such technology and ensure it is leveraged for beneficial purposes. 
\clearpage  
\clearpage  

%
%
\section*{Acknowledgment}
This work was supported in part by the National Natural Science Foundation of China under Grant 62125109, Grant 61931023, Grant 61932022, Grant 62371288, Grant 62320106003, Grant 62301299, Grant T2122024, Grant 62120106007.
\bibliographystyle{splncs04}
\bibliography{main}

\appendix

\section{More Implementation Details}
\noindent\textbf{Dataset.}
For multi-object segmentation in video, we evaluate our method on one synthetic and two real-world video datasets.
\textbf{MOVi-E}~\cite{greff2021kubric} dataset is a synthetic dataset with granular control over data complexity and comprehensive ground truth annotations. MOVi-E scenes contain up to 23 objects and introduce simple linear camera movement.
Our evaluation of learned features extends to real-world datasets \textbf{DAVIS-17}~\cite{pont20172017} and \textbf{YouTube-VIS-19}~\cite{Yang2019vis}. DAVIS-17, an expansion of DAVIS-16, includes 40 additional video sequences along with multi-object segmentation annotations. We utilize 30 validation videos on DAVIS-2017 for evaluation. As for YouTube-VIS-19, due to the lack of mask annotations in the validation or test set, following previous works~\cite{aydemir2023self,zadaianchuk2023objectcentric}, we select 300 out of the whole 2,883 videos in the training set for evaluation. For MOVi-E and YouTube-VIS-19, we report the Foreground Adjusted Rand Index (FG-ARI) and mean Intersection over Union (mIoU). Furthermore, for DAVIS-2017, we adhere to the standard protocol~\cite{pont20172017} and report both Region Similarity ($\mathcal{J}$) and Contour Accuracy ($\mathcal{F}$).

\noindent\textbf{Hierarchical Clustering Algorithm.} We present our Hierarchical Clustering based inference in Alg.~\ref{alg:cluster}. Given the spatio-temporal attention maps $A_v\in\mathbb{R}^{THW\times THW}$, the algorithm finally outputs the cluster centers $A_c\in\mathbb{R}^{N\times THW}$ and cluster assignments $Z\in\{1,...,N\}^{T'HW}$ which serve as predicted object segmentation masks. 
Specifically, each attention map is treated as a separate cluster. The process then cycles through each attention map (or current `cluster'), calculating distances between it and all other clusters using the KL-divergence metric. It identifies the clusters that are close to it (i.e., those whose distance is less than the threshold) and combines them to form a new, larger cluster, represented by their updated centroid. This updated cluster set then replaces the initial set of attention maps, and the process continues iteratively until no more clusters can be merged.
Finally, the algorithm assigns each original attention map to the cluster whose center it is closest to, yielding the final cluster assignments.
Note that executing inference on extensive video sequences with a large $T$ value might cause the self-attention matrix to become redundant, thereby requiring significant computational resources. To address this limitation, we sparsely sample $T'$ frames ($T'<<T$) as \texttt{key}, with the original densely sampled frames as \texttt{query}, and calculate the cross-attention $A'_v\in\mathbb{R}^{THW\times T'HW}$. By applying clustering to more compact $A'_v$, we linearly reduce memory requirements and maintain stable performance as shown in Sec.~\ref{samplenum}. 


\begin{algorithm*}[]
\small
\caption{Hierarchical Clustering}
\label{alg:cluster}
\begin{algorithmic}
\State\textbf{Input:} Spatio-temporal attention maps $A_v\in\mathbb{R}^{THW\times T'HW}$, distance threshold $\tau$
\State\textbf{Output:} Cluster assignment $Z \in \{1,...,N\}^{THW}$
\State Initialize cluster centers $A_c \leftarrow A_v$
\While{the number of clusters in $A_c$ changes}
    \State Initialize updated clusters $A_p \leftarrow \{\}$
    \ForAll{$x\in A_c$}
        \State Compute distances: $\mathcal{M} \leftarrow \texttt{calculate\_distance}(x,A_c)$
        \State Identify proximal members: $\mathcal{I} \leftarrow \{i \mid \mathcal{M}[i]<\tau\}$
        \State Calculate new cluster centroid: $x \leftarrow \frac{1}{|\mathcal{I}|}\sum_{i\in\mathcal{I}}A_c[i]$ 
        \State Add new centroid to updated clusters: $A_p \leftarrow A_p\cup \{x\}$
        \State Remove merged attention maps from current set: $A_c \leftarrow A_c\backslash A_c[i] , \forall i\in\mathcal{I}$ 
    \EndFor
    \State Update the clusters: $A_c \leftarrow A_p$ 
\EndWhile
\State Compute final distances: $\mathcal{M} \leftarrow \texttt{calculate\_distance}(A_v,A_c)$
\State Compute final cluster assignments: $Z \leftarrow \texttt{argmin}(\mathcal{M},\texttt{dim=1})$ 
\end{algorithmic}
\end{algorithm*}

\section{Unsupervised Single Object Segmentation}
Besides multi-object segmentation, our method also works in single-object scenarios. We benchmark on three popular datasets designed for single-object segmentation. 
\textbf{DAVIS-16}~\cite{perazzi2016benchmark} consists of 50 high-quality videos, 3455 frames in total. Every frame is annotated with a pixel-level accurate segmentation mask. 
\textbf{SegTrack-v2}~\cite{li2013video} contains 14 sequences and 947 fully-annotated frames. Each sequence involves 1-6 moving objects and presents challenges including motion blur, appearance change, complex deformation, occlusion, slow motion, and interacting objects.
\textbf{FBMS-59}~\cite{ochs2013segmentation} has 59 sequences with greatly varied resolution and annotates every 20th frame. Many sequences contain multiple moving objects. Following previous evaluation metric~\cite{Yang_2019_CVPR,xie2022segmenting}, we merge objects of SegTrackv2 and FBMS-59 into a single one for video object segmentation.
We calculate the mean per-frame the Jaccard Index 
$\mathcal{J}$ over the validation set.
In single-object segmentation benchmarks that annotate all objects collectively, we set the distance threshold to 1.6 to combine all foreground objects into one cluster. 
\begin{table}[]
\small
    \centering
    \caption{Quantitative results on single object video segmentation. The tick($\checkmark$) and cross(\xmark) labels under the RGB and Flow columns indicate whether a method utilizes the corresponding modality during training or inference. We compare per frame mean IoU on DAVIS-16, SegTrack-v2 and FBMS-59 without any post-processing (e.g., spectral clustering, test-time adaptation, CRF~\cite{lafferty2001conditional}).}
    \begin{tabular}{lccccc}
    \toprule
        Model & RGB & Flow & DAVIS & ST-v2 & FBMS \\
        \midrule
        NLC~\cite{faktor2014video} & \checkmark & \checkmark & 55.1 & 67.2& 51.5\\
        CIS~\cite{Yang_2019_CVPR} & \checkmark & \checkmark & 59.2 & 45.6 & 36.8 \\        
        TokenCut~\cite{wang2022tokencut2} & \checkmark & \checkmark &  64.3 & 60.2 & 59.6 \\
        DyStaB~\cite{Yang_2021_CVPR} & \checkmark & \checkmark & 80.0 & 74.2 & 73.2   \\
        DeSprite~\cite{ye2022deformable} & \checkmark & \checkmark & 79.1 & 72.1 &  71.8 \\
        RCF~\cite{lian2023bootstrapping} & \checkmark & \checkmark & 80.9 & 76.7 & 69.9 \\
        SIMO~\cite{lamdouar2021segmenting} & \xmark & \checkmark & 67.8 & 62.0 & - \\
        MG~\cite{Yang_2021_ICCV} & \xmark & \checkmark & 68.3 & 58.6 & 53.1 \\
        EM~\cite{meunier2022driven} & \xmark & \checkmark & 69.3 & 55.5 & 57.8 \\
        OCLR~\cite{xie2022segmenting} & \xmark & \checkmark & 72.1 & 67.6 & 65.4 \\
        \midrule
        AMD~\cite{liu2021emergence} & \checkmark & \xmark & 57.8 & 57.0 & 47.5 \\        
        SMTC~\cite{qian2023semantics} & \checkmark & \xmark & 71.8 & 69.3 & 68.4 \\
        \textbf{Ours} & \checkmark & \xmark & 75.4 & 74.8 & 73.3  \\
        \bottomrule
    \end{tabular}
    \label{tab:single}
\end{table}

We present the quantitative results on unsupervised single object discovery in Table~\ref{tab:single}. Note that all the compared methods are trained on the target dataset, while our model is only trained on YouTube-VIS-19 and directly transferred to these single object segmentation benchmarks in a \emph{zero-shot} manner. Despite this, our method still achieves the best performance among those only using RGB data. 
Though SMTC~\cite{qian2023semantics} proposes a sophisticated VOS framework based on slot attention, our method outperforms it by approximately 5 points.
The superiority demonstrates the generalization ability of our approach simply guided by attention. As for the counterparts that resort to optical flow, some of them achieve very promising performance on three benchmarks
~\cite{yang2021dystab,lian2023bootstrapping,ye2022deformable}. 
It is because optical flow strongly prioritizes moving areas in videos, making it particularly well-suited for single object segmentation tasks. However, the utility of optical flow may diminish for multi-object setups. For instance, it becomes complicated to distinguish between two objects moving in the same direction based solely on flow information. Moreover, it can be difficult to obtain a reliable flow in complex scenarios.
Conversely, our method eliminates the need for any optical prior and can be conveniently adapted to accommodate multi-object scenarios.

\section{More Ablation Study}

\begin{table}[]
    \centering
    \small
    \caption{Ablation on different pretrained backbones. We show the results on various DINO and DINOv2 pretrained ViT encoders with different patch sizes.}
    \begin{tabular}{lcccc}
    \toprule
        {} & \multicolumn{2}{c}{YTVIS-19} & \multicolumn{2}{c}{DAVIS-17} \\
        \cmidrule(r){2-3}
        \cmidrule(r){4-5}
        Model & FG-ARI & mIoU & $\mathcal{J}\&\mathcal{F}$ & FG-ARI \\
        \midrule
        DINO ViT-S/16 & 42.5 & 47.2 & 41.7 & 38.3 \\
        DINO ViT-S/8 & 44.3 & 50.1 & \textbf{43.9} & 40.1 \\
        DINO ViT-B/8 & 43.5 & \textbf{50.2} & 42.8 & 40.7 \\
        DINOv2 ViT-S/14 & 44.1 & 49.7 & 43.1 & 40.5 \\
        DINOv2 ViT-B/14 & \textbf{44.5} & 50.1 & 43.7 & \textbf{41.6} \\
    \bottomrule
    \end{tabular}
    \label{backbone}
\end{table}
\noindent\textbf{Pretrained backbones.}
We present the ablation studies on different pretrained backbones in Table.~\ref{backbone}. We show the results on both DINO and DINOv2 pretrained ViT encoders with different patch sizes. Generally, our method achieves competitive results on all variants of visual encoders. Comparing the first two lines, i.e., DINO ViT-S/16 vs. DINO ViT-S/8, smaller patch size contributes to notable performance improvements, approximately 2 points on two benchmarks, due to more fine-grained segmentation predictions. Comparing DINO and DINOv2, despite larger patch size, more advanced DINOv2 pretrained backbones reach comparable performance. This reveals the robustness and flexibility of our method to different backbones.

\begin{table}[]
    \centering
    \small
    \caption{Ablation on different numbers of key frames sampled for calculating the spatio-temporal attention matrix. We compare performance under different ratios.}
    \begin{tabular}{cccccc}
    \toprule
        {} & \multicolumn{2}{c}{YTVIS-19} & \multicolumn{2}{c}{DAVIS-17} & Speed \\
        \cmidrule(r){2-3}
        \cmidrule(r){4-5}
        \cmidrule(r){6-6}
        Ratio & FG-ARI & mIoU & $\mathcal{J}\&\mathcal{F}$ & FG-ARI & Ratio \\
        \midrule
        0.1 & 43.5 & 49.7 & 42.8 & 40.0 & 2.3$\times$ \\
        0.2 & 44.1 & 50.0 & 43.4 & 40.2 & 2.0$\times$ \\
        0.5 & 44.3 & 50.2 & 43.8 & 40.1 & 1.4$\times$ \\
        1.0 & 44.3 & 50.1 & 43.9 & 40.1 & 1.0$\times$ \\
    \bottomrule
    \end{tabular}
    \label{sample}
\end{table}

\noindent\textbf{Number of key frames.}
\label{samplenum}
As stated in the above section, it is feasible to sparsely sample video frames as \texttt{key} to reduce computation costs in inference. We present the ablation study in Table~\ref{sample}. We report the multiple object segmentation performance as well as the inference throughput ratio (including the whole feature extraction, attention calculation and clustering process). Interestingly, our findings suggest that promising results for video object segmentation can still be achieved even when only 10\% of the frames are sampled. This sparse sampling approach leads to a remarkable $2.3\times$
speedup in inference. For illustration, given a video with $T$ frames, we uniformly sample $T'=0.1T$ frames as \texttt{key}, with the original $T$ frames as \texttt{query}, and calculate the cross-attention $A'_v\in\mathbb{R}^{THW\times T'HW}$. This linearly reduces the channel dimension of each attention map. Then we perform hierarchical clustering on these $THW$ samples with reduced channel dimension and produce the cluster assignments (segmentation masks) for all frames within the video in one shot. The underlying reason is that video frames are highly redundant, sparse sampling could provide an abundant temporal reference for spatio-temporal dependency calculation. Hence, by sampling a small percentage of the entire video, we achieve comparable performance with a substantial reduction in computational cost, leading to faster inference speeds.
Furthermore, techniques such as quantization~\cite{li2024frequencyaware,ding2024ampa,lu2024terdit}  and pruning~\cite{ding2023prune,lu2024spp,lu2024not,NEURIPS2023_62c9aa4d} can provide additional speedups for the method.

\begin{figure}
    \centering
    \subfloat[Results on motocross-jump sequence.]{\includegraphics[width=0.5\linewidth]{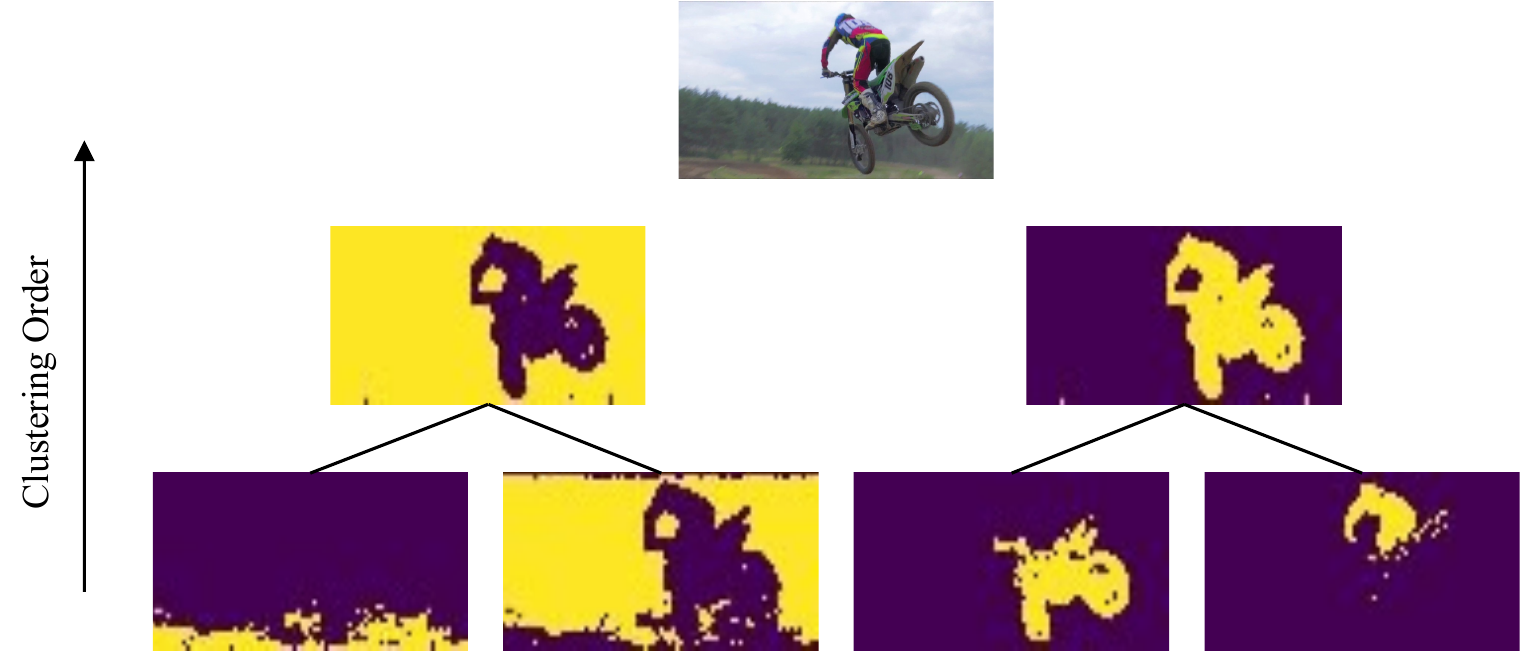}
    \label{motohuman}} \\
    \subfloat[Results on parkour sequence.]{\includegraphics[width=\linewidth]{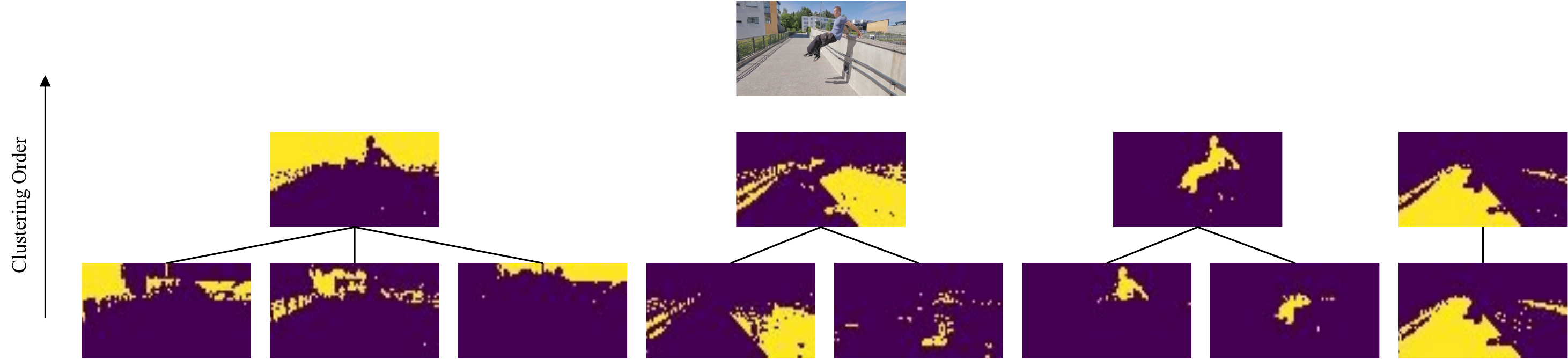}
    \label{parkour}}
    \caption{Visualization of the clustering process. We observe interpretable clustering hierarchies that segment objects at different granularities.}
    \label{hierarchy}
\end{figure}
\section{More Visualization Results}
\noindent\textbf{Results of different clustering hierarchies.} In our inference stage, the hierarchical clustering algorithm produces different clustering hierarchies. We explore whether there exists an interpretable phenomenon during the clustering process in Fig.~\ref{hierarchy}. 

In each hierarchy, we gather the centroids within a distance threshold $\tau$ into a new centroid that covers a larger area. Generally, as clustering goes, the hierarchy increases and the result transitions from fine-grained to coarse-grained. For example, the lower hierarchy results in human body parts, and the higher hierarchy results in the whole foreground object like human and motorbike. Smaller $\tau$ will force the clustering to stop at lower hierarchies, thus generating fine-grained segmentation. Larger $\tau$ will continue to merge the fine-grained cluster centroids, e.g., merge human body parts into a whole human. We find $\tau = 1.0$ works empirically well for various benchmarks.

Interestingly, we observe that our model is able to segment objects at different granularities across hierarchies. Generally, it results in more fine-grained object segmentation in lower hierarchies and vice versa. For example, in Fig.~\ref{motohuman}, the model discerns two distinct objects - the motorbike and the human - at a lower hierarchy, subsequently merging them into a cohesive foreground area at a higher hierarchy. Similarly, Fig.~\ref{parkour} shows that the clustering isolates different sections of the human figure at a lower hierarchy, before integrating them to form a holistic human body at a higher hierarchy. Such interpretable hierarchical clustering outcomes yield multi-layered object segmentations, potentially resolving the ambiguities in annotations.

\begin{figure*}
    \centering
    \subfloat[Results on cat-girl sequence.]        {\includegraphics[width=\linewidth]{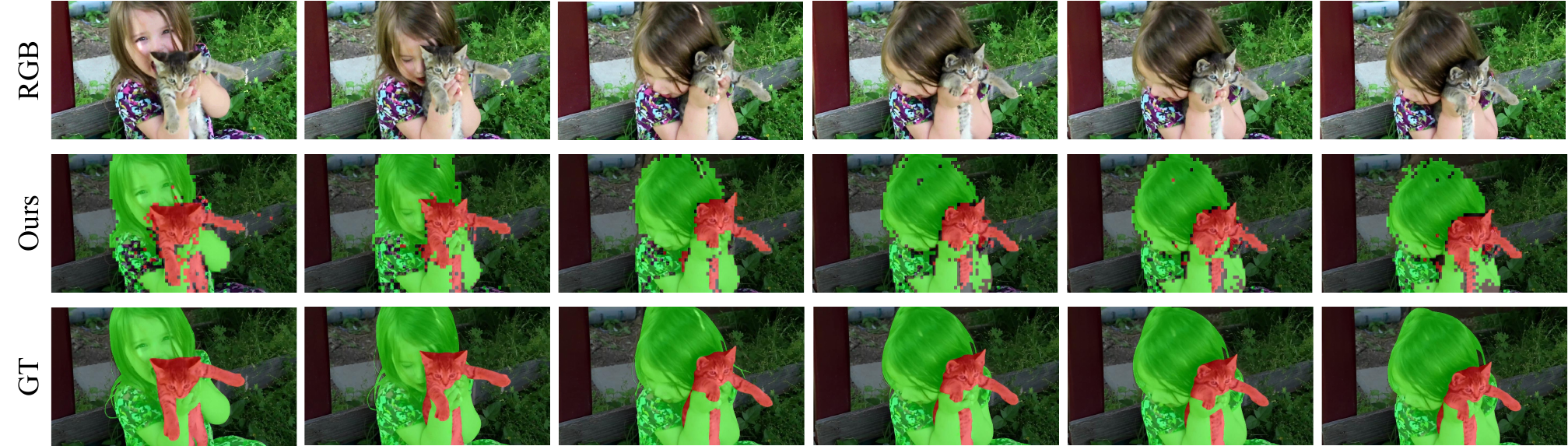}}\\
    \subfloat[Results on kid-football sequence.]        {\includegraphics[width=\linewidth]{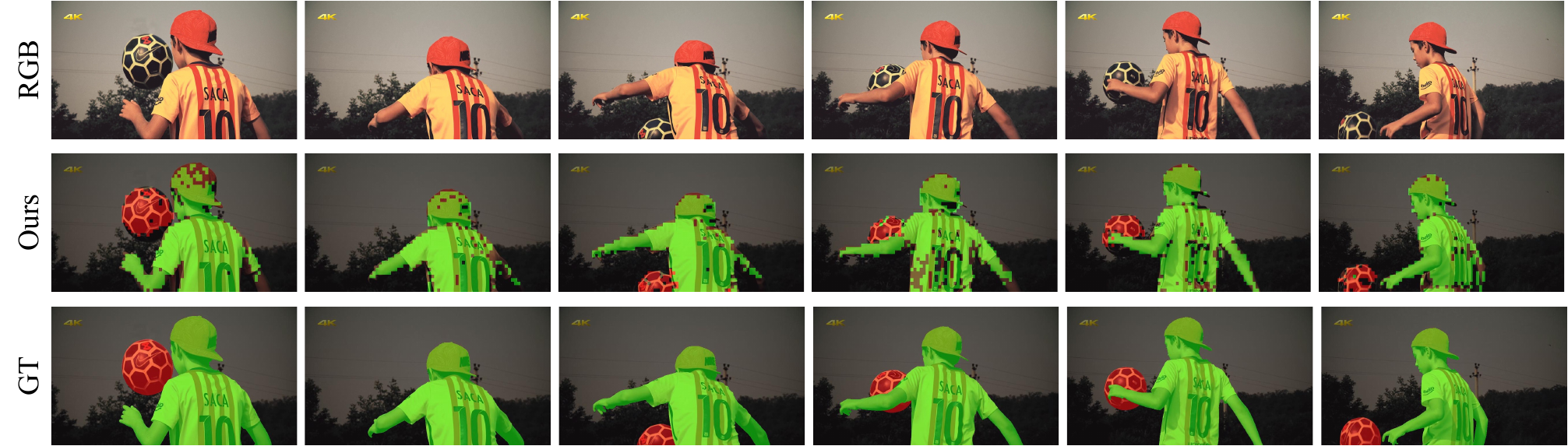}}\\
    \subfloat[Results on lizard sequence.]        {\includegraphics[width=\linewidth]{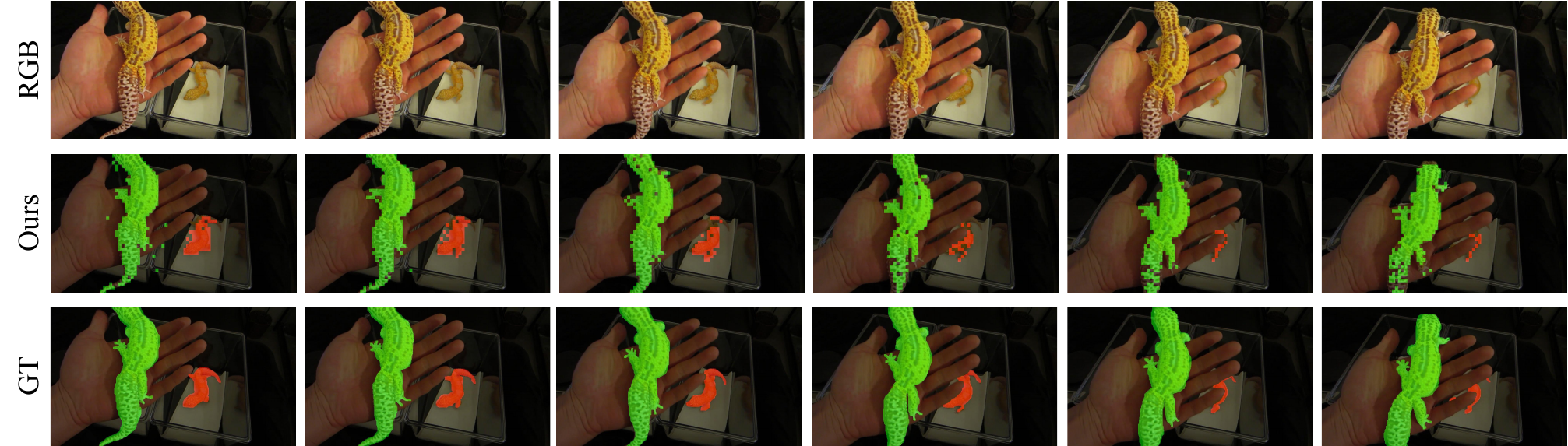}}
    \caption{Visualization results on video sequences with occlusion. Our model is able to deal with partial or complete object occlusion, where an object disappears in some frames and reappears in later frames.}
    \label{occlusion}
\end{figure*}

\noindent\textbf{Results on consecutive frames.} We additionally show our segmentation results on video sequences with object occlusion, disappearance and reappearance, which is prevalent and challenging in real-world scenarios. In Fig.~\ref{occlusion}, we present three typical cases. The first is a cat-girl sequence, where there exist mutual occlusions between two objects. Our model is able to accurately segment the object parts despite severe occlusion. The second is a kid-football sequence, where the football disappears in the second frame and reappears in later frames. Since our method refers to the spatio-temporal dependencies across the whole temporal range, it is able to recognize that the ball in the first frame and those in later frames belong to the same instance. This enables our model to process real-world videos with complex temporal dynamics. The third is a very challenging sequence consisting of two lizards, which share very similar colors, body shapes, and textures and only vary in sizes and positions. Moreover, the smaller one is severely occluded by the human hand in the latter three frames. Despite these challenges, our method is still able to distinguish these two lizards and accurately track specific instances over time. These examples demonstrate the applicability of our method to general video scenes.

\end{document}